\pdfoutput=1

\documentclass[11pt]{article}

\usepackage{latex/acl}
\usepackage{graphicx}
\usepackage{times}
\usepackage{latexsym}

\usepackage{algorithm}
\usepackage{algorithmic}
\usepackage{hyperref}
\usepackage{url}
\usepackage{CJKutf8}
\usepackage{bigstrut}
\usepackage[normalem]{ulem}
\usepackage{amssymb}
\usepackage{amsfonts}
\usepackage{multicol}
\usepackage{tipa}
\usepackage{amsmath}
\usepackage{bm}
\usepackage{tikz}
\usepackage{paralist}
\usepackage{IEEEtrantools}
\usepackage[switch]{lineno}
\usepackage{enumitem}
\usepackage{multirow, makecell, caption}
\usepackage{arydshln}
\usepackage{verbatim}
\usepackage{amssymb}
\usepackage{pifont}
\usepackage{csquotes}
\usepackage{xspace}
\usepackage{mdwlist}
\usepackage{subfigure}
\usepackage{array}
\usepackage{colortbl}
\usepackage{dsfont}
\usepackage{bibentry}
\usepackage{tabularx}
\usepackage{bbm}
\usepackage{pgfplots}
\usepackage{soul} 
\usepackage{color,xcolor} 
\usepackage{booktabs}
\usepackage[T1]{fontenc}
\usepackage[utf8]{inputenc}
\usepackage{microtype}
\usepackage{inconsolata}

\newcommand{\ie}{\textit{i.e.}\xspace}
\newcommand{\eg}{\textit{e.g.}\xspace}

\newcommand{\method}{\textsc{COG}\xspace}
\newcommand{\CI}{\textsc{Concept Integration}\xspace}
\newcommand{\EF}{\textsc{Evidence Fusion}\xspace}

\title{Light Up the Shadows: Enhance Long-Tailed Entity Grounding with Concept-Guided Vision-Language Models}

\author{Yikai Zhang\textsuperscript{\rm $\spadesuit$},
Qianyu He\textsuperscript{\rm $\spadesuit$},
Xintao Wang\textsuperscript{\rm $\spadesuit$},\\
\bf Siyu Yuan\textsuperscript{\rm $\heartsuit$},
Jiaqing Liang\textsuperscript{\rm $\heartsuit$} \thanks{~~Corresponding authors.},
Yanghua Xiao\textsuperscript{\rm $\spadesuit$}\footnotemark[1]
\\
\textsuperscript{\rm $\spadesuit$}Shanghai Key Laboratory of Data Science, School of Computer Science, Fudan University\\
\textsuperscript{\rm $\heartsuit$}School of Data Science, Fudan University\\
\texttt{\{ykzhang22,qyhe21,xtwang21,syyuan21\}@m.fudan.edu.cn},\\
\texttt{\{liangjiaqing, shawyh\}@fudan.edu.cn}}

\pgfplotsset{compat=1.18}
\begin{document}
\maketitle
\begin{abstract}
Multi-Modal Knowledge Graphs (MMKGs) have proven valuable for various downstream tasks. 
However, scaling them up is challenging because building large-scale MMKGs often introduces mismatched images (\ie, noise). 
Most entities in KGs belong to the long tail, meaning there are few images of them available online. 
This scarcity makes it difficult to determine whether a found image matches the entity. 
To address this, we draw on the Triangle of Reference Theory and suggest enhancing vision-language models with concept guidance. 
Specifically, we introduce \method, a two-stage framework with \textbf{CO}ncept-\textbf{G}uided vision-language models. 
The framework comprises a \CI module, which effectively identifies image-text pairs of long-tailed entities, and an \EF module, which offers explainability and enables human verification.
To demonstrate the effectiveness of \method, we create a dataset of 25k image-text pairs of long-tailed entities. 
Our comprehensive experiments show that \method not only improves the accuracy of recognizing long-tailed image-text pairs compared to baselines but also offers flexibility and explainability.\footnote{Resources of this paper can be found at \url{https://github.com/ykzhang721/COG}.} 
\end{abstract}

\section{Introduction}
\label{sec:intro}
Multi-Modal Knowledge Graphs (MMKGs) are knowledge graphs that integrate and align information from diverse modalities (\eg, text and images)~\cite{ferrada2017imgpedia,liu2019mmkg,wang2020richpedia}.
Due to the growing demand for multi-modal intelligence and extensive knowledge in various applications~\cite{hou2019relational,marino2021krisp}, MMKGs have received increasing attention in recent years. 

\begin{figure}[t] 
    \centering
        \includegraphics[width=1.0\linewidth]{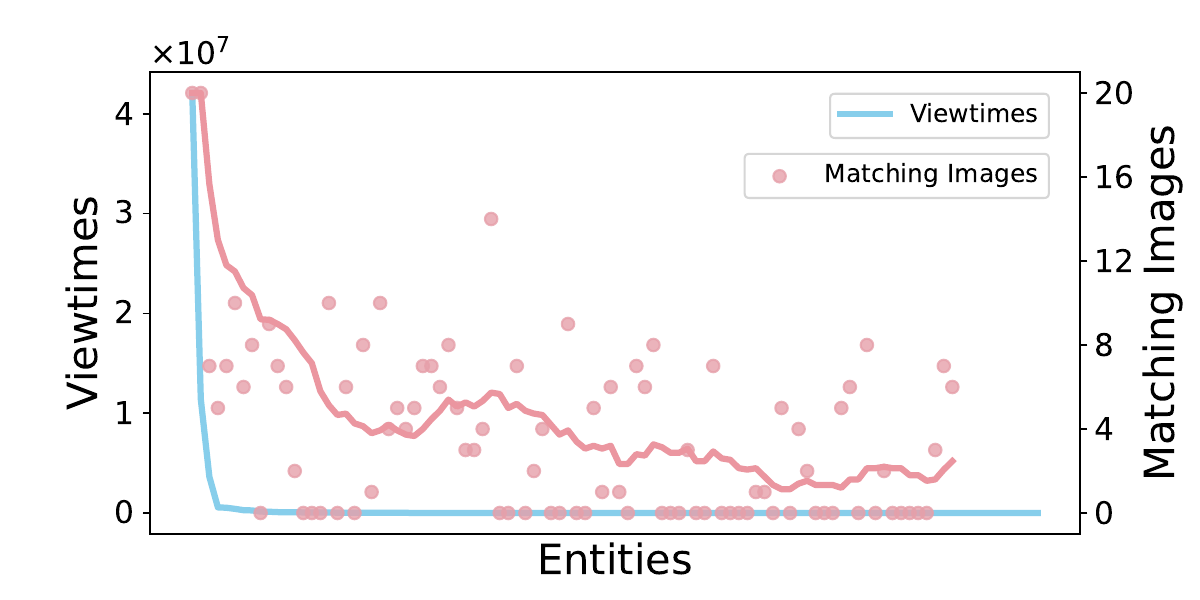} 
    \caption{
    We randomly select 100 entities from the large-scale knowledge graph CN-DBpedia~\cite{xu2017cn} and add human annotations. 
    The blue line represents the changes in the entities' \textit{viewtimes}, which indicates their click frequency. 
    The red dots indicate the number of correctly matched images found in the top 20 search results for each entity, and the red line smooths out these data points.
    }
    \label{fig:powerlaw}
\end{figure}

Although the number of images in current MMKGs has increased, their coverage and accuracy are limited~\cite{onoro2017answering,wang2020richpedia}, especially for long-tailed entities (\ie, less common entities). 
Figure~\ref{fig:powerlaw} illustrates that the trends in an entity's click frequency and the number of matching images found are similar, both displaying a long-tailed pattern, indicating the scarcity of images of long-tailed entities.

Aligning long-tailed entities with appropriate images (\ie, entity grounding) is crucial in constructing MMKGs. 
First, it expands the scope and completeness of MMKGs, which traditionally focus on common entities~\cite{onoro2017answering,wang2020richpedia}. 
Second, adding visual content for long-tailed entities boosts efficiency in downstream tasks~\cite{pezeshkpour2018embedding,chen2022re,chen2022murag}. 
Third, pairing images with long-tailed entities provide valuable training data for developing and refining vision-language models, particularly for rare or domain-specific entities.

\begin{figure}[t] 
    \centering
        \includegraphics[width=0.8\linewidth]{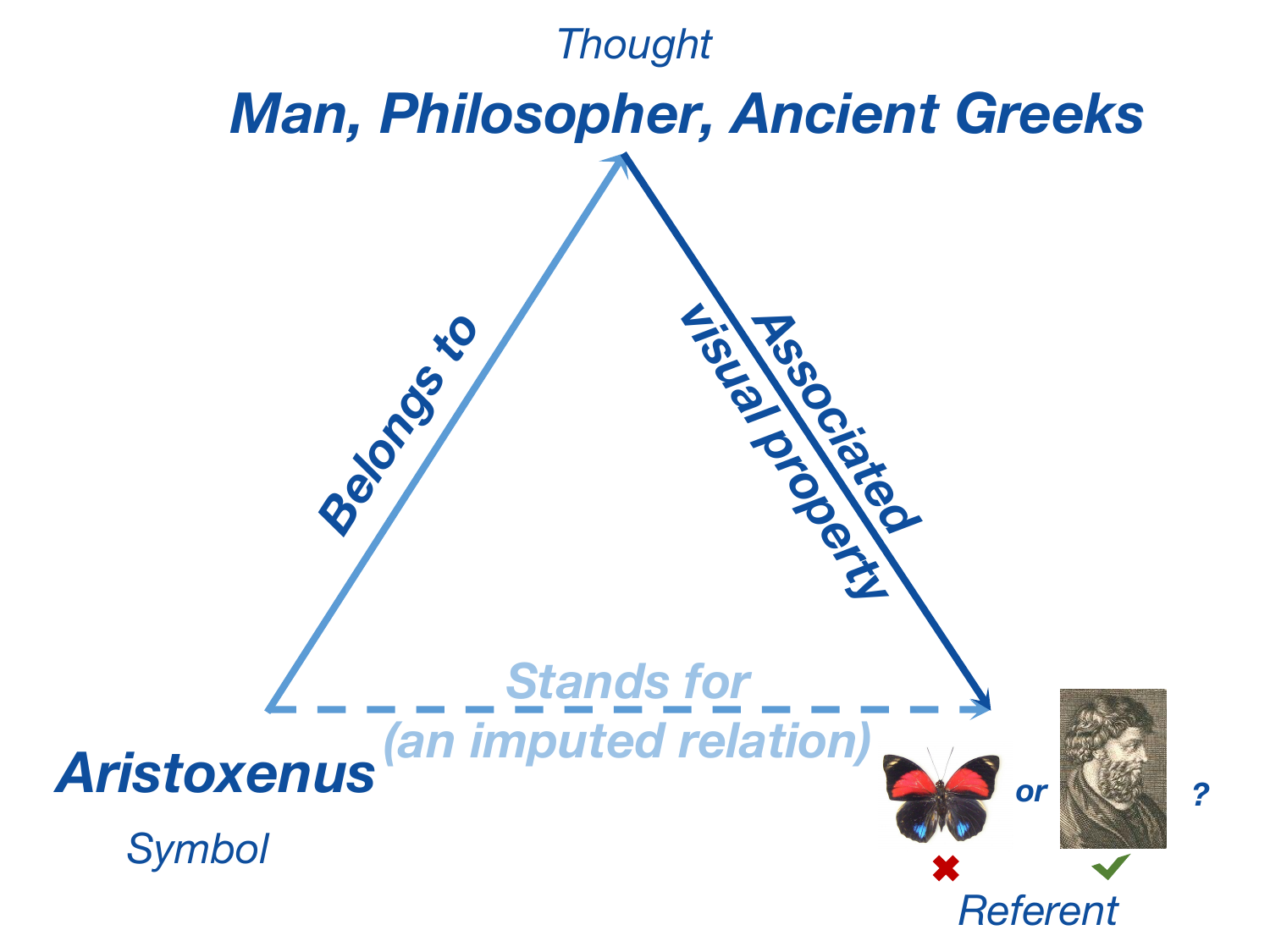} 
    \caption{
    This figure shows that when searching for an entity named \textit{Aristoxenus}, the search engine returns two images. By applying concepts, we can conclude that the target \textit{Aristoxenus} refers to a person, not a butterfly.}
    \label{fig:triangle}
\end{figure}

Grounding long-tailed entities in MMKGs is challenging. 
Current methods~\cite{onoro2017answering,liu2019mmkg,wang2020richpedia} for entity grounding rely on web resources, especially search engines.
They collect images by matching image captions to entity names. 
While effective for well-known entities~\cite{liu2019mmkg,wang2020richpedia}, these approaches struggle with long-tailed entities due to several limitations:
\begin{inparaenum}[1)]
    \item Search engines used for text matching show suboptimal performance because entity grounding involves matching images and texts;
    \item Although pre-trained vision-language models (PVLMs) have shown impressive performance in various cross-modal tasks, they encounter challenges in recognizing long-tailed entities due to their infrequent appearance during pre-training;
    \item Current methods lack the ability to explain their image selection, which is important for ensuring quality.
\end{inparaenum}

To address these challenges, we develop \method, a two-stage framework that uses \textbf{CO}ncept-\textbf{G}uided vision-language models to ground long-tailed entities. 
Initially, as shown in Figure~\ref{fig:triangle}, the Triangle of Reference Theory~\cite{mcelvenny2014ogden} explains the connection between \textit{Thought}, \textit{Symbol}, and \textit{Referent}. 
This demonstrates how humans link an entity to a real-world object through concepts. 
Inspired by this theory, we improve PVLMs to accurately identify images of long-tailed entities using concepts. 
Notably, our \method allows for the use of replaceable PVLMs, meaning it can integrate any module for matching images and texts, thus offering flexibility in application. 
Next, we investigate how the choice of concept selection strategy affects the performance of \method, considering the varying scope of concepts. 
Lastly, \method not only enhances recognition accuracy but also ensures explainability, providing a basis for further quality control (\ie, human verification).

To sum up, our contributions are as follows:
\begin{itemize}
    \item We introduce a flexible and explainable two-stage framework \method, using concept-guided PVLMs to recognize image-text pairs of long-tailed entities. 
    \item We examine and discuss how choosing different concepts affects the experimental results.
    \item Through extensive experiments, we show that our method significantly enhances the accuracy of long-tailed entity grounding. It also allows for human verification, ensuring finer quality control.
\end{itemize}

\section{Related Work}
\label{sec:related works}
\subsection{Multi-Modal Knowledge Graph Construction}
A Multi-Modal Knowledge Graph (MMKG) is a unified information representation that integrates data from various modalities, such as text, images, and audio, into a single interconnected graph~\cite{zhu2022multi}. 
Existing methods for entity grounding in MMKGs can be categorized into two main groups:
\begin{inparaenum}[\it 1)]
\item \textit{Methods based on online encyclopedias}~\cite{ferrada2017imgpedia,alberts2020visualsem}: These methods link existing encyclopedic multimedia resources (\eg, Wikimedia Commons, Wikipedia, ImageNet~\cite{deng2009imagenet}) by associating texts with images to construct MMKGs;
\item \textit{Methods based on web search engines}~\cite{onoro2017answering,liu2019mmkg,wang2020richpedia}: These methods directly search for images of entities using web search engines. 
This approach is more flexible than using online encyclopedic multimedia data, as it allows for expansion based on existing filtered and refined KGs. 
However, it tends to prioritize popular entities because entity images adhere to a power-law distribution. 
For long-tailed entities lacking web images, search engines, despite providing ranked results, can easily return incorrect or mismatched images, leading to noise.
\end{inparaenum}
In this paper, we propose \method that leverages concepts to reduce this kind of noise for long-tailed entity grounding.

\subsection{Pre-Trained Vision-Language Models}
Pre-trained Vision-Language Models (PVLMs) are designed to process visual and textual data, aiming to align image-text data through extensive cross-modal pre-training. 
Many approaches conduct contrastive pre-training on large-scale datasets~\cite{radford2021learning,jia2021scaling,yuan2021florence,li2022blip,yu2022coca}. 
CLIP~\cite{radford2021learning} employs a self-supervised method with 400 million internet-sourced image-text pairs to enhance modality alignment. 
ALIGN~\cite{jia2021scaling} uses a dual-encoder and trains with over a billion pairs, while BLIP~\cite{li2022blip} improves multi-modal task performance by filtering out low-quality data. 
In \method, we design a two-stage framework to guide PVLMs with concepts. 
PVLMs serve as a module to generate similarity between texts and images in \method. 
This module can be replaced by other methods with similar functionality, making our framework more flexible and general.

\subsection{Long-Tailed Classification}
Some researchers in computer vision focus on the problem of long-tailed image classification. 
This issue primarily concerns classifying images that are infrequent in the training set. 
Various datasets~\cite{liu2019large,cui2019class} are used to evaluate the ability to learn classification with limited samples. 
However, the problem we discuss in this paper is different from traditional image classification. 
Long-tailed entity grounding is not confined to known, fixed classes, which still include numerous images on the web. 
Identifying images of specific and unknown entities poses a significant challenge to PVLMs. 
To address this issue, we suggest creating connections between entities and images for PVLMs using concepts, aiming for accurate recognition.

\section{\method}
\label{sec:Method}
\subsection{Problem Definition}
A Multi-Modal Knowledge Graph (MMKG) is a knowledge graph where nodes are entities or images, and edges represent their relationships. 
In MMKG, triplets are defined as $(e, \textit{has image}, i)$, where $e$ is the textual entity and $i$ is its corresponding image, indicating a \textit{has image} relationship.
To match images with entities in MMKG (\ie, entity grounding in MMKG), common methods typically follow a two-step process. 
First, they rank the collected images based on their relevance to the given entity, which can be modeled as a \textbf{Ranking} task. 
To formalize this, given a corrupted triplet 
$(e, \textit{has image}, ?)$
in MMKG, this sub-task aims to predict the removed image $i$. 
Then, they select the top-$n$ images and classify whether the image is related to the given entity in order, which can be modeled as a binary \textbf{Classification} task. 
To formalize this, each triplet 
$(e, \textit{has image}, i)$
can be classified as $True$ if the image correctly matches the entity; otherwise, the triplet is classified as $False$. 
We design both tasks to thoroughly simulate the entity grounding process and conduct rich experiments on \method.

\subsection{Concept Selection}
A concept typically refers to a group of entities sharing common characteristics. 
These concepts can be categorized based on the number of entities they encompass, indicating different levels of granularity. 
To identify which concepts are most effective for concept guidance, we explore the influence of utilizing various concepts. 
It is common for entities to embody multiple concepts, each displaying distinct levels of granularity. 
According to \cite{wang2015inference}, humans mainly utilize Basic-level Categorization (BLC), a mid-level concept, for everyday thinking.

Motivated by this understanding, we evaluate the efficiency of BLC concepts compared to a broader range of concepts. 
We describe BLC concepts as those represented by a single word and examine their performance against that of all concepts. 
The experiments outlined in $\mathsection$~\ref{exp:analysis} demonstrate how different strategies for selecting concepts influence performance.

\subsection{Details of \method}

\begin{figure*}[t]
    \centering
        \includegraphics[width=1.0\linewidth]{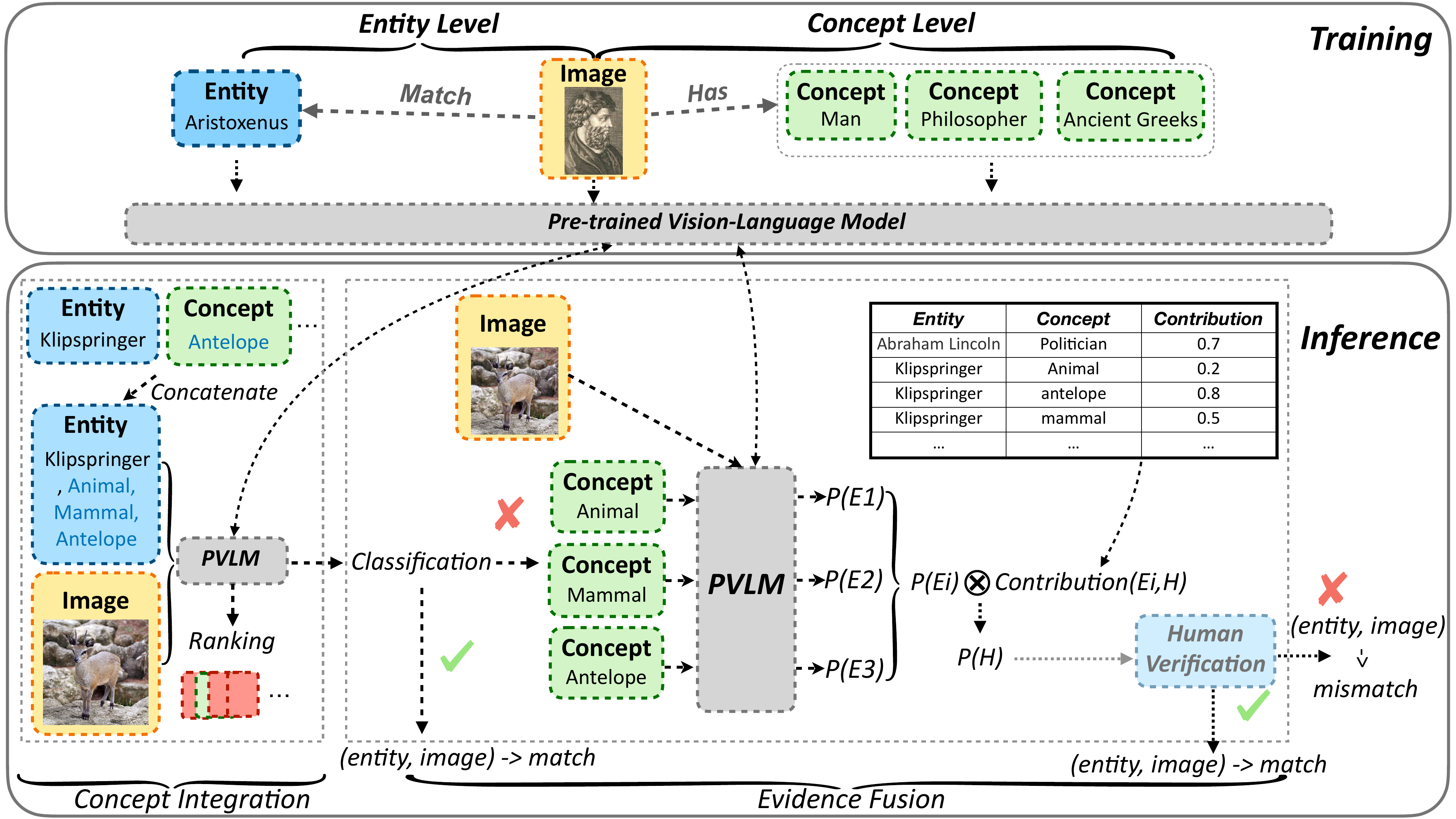} 
    \caption{Overview of \method. \method uses contrastive learning on entity and concept levels for model training. At the inference stage, we utilize a two-stage framework with \CI and \EF modules. \CI aims for direct prediction of image-text matches using concept guidance, while the \EF module reassesses discarded image candidates from \CI, particularly valuable for rare, long-tailed entities.}
    \label{fig:framework}
\end{figure*}

Concept guidance initially necessitates that PVLMs possess the capability to identify concepts and images. Subsequently, they should use this ability to conduct thorough analyses for long-tailed entity grounding.

Figure~\ref{fig:framework} illustrates that, to train models for concept recognition, we use contrastive learning on both the entity and concept levels, as detailed in $\mathsection$~\ref{method:train}.
When performing inference with our finely tuned model, we introduce a two-stage framework consisting of two modules in $\mathsection$~\ref{method:inference}: \CI and \EF. 
\CI combines all available information to directly predict whether an image matches the corresponding text. 
Given that images of long-tailed entities are rare and valuable, we develop \EF to reassess the image candidates discarded by \CI. 
\EF offers clear evidence by breaking down various attributes of the entity, thus providing a basis for human annotation.

\subsubsection{Contrastive Learning on Two Levels}
\label{method:train}
When training PVLMs, we designate a text as $t$ and an image as $i$.
First, we input both $t$ and $i$ into the PVLM. 
The model generates a $prediction$ that indicates the degree of alignment between $t$ and $i$, as demonstrated here:
\begin{align}
    logit &= PVLM(t, i) \\
    Sigmoid(logit) &= \frac{1}{1 + e^{-logit}} \\
    prediction &= Sigmoid(logit)
\end{align}

In this equation, both $t$ and $i$ symbolize the text and image inputs respectively. 
The $prediction$ value signifies the model's prediction of the similarity between the image and the text. 
If the $prediction$ surpasses a certain threshold, we deem it as a match; otherwise, it is considered a mismatch.

Next, we train the model using contrastive learning, which includes in-batch negative samples. 
Each batch contains $n$ samples, and $n$ represents the batch size. 
A sample consists of a pair $(t, i)$, which stands for a text and an image. 
As shown in Figure~\ref{fig:framework}, we create contrastive samples on both the entity and concept levels. 
We define \(t_i\) as the combination of the \(i\)-th entity and its concepts: \(t_i = e_i,c_1,c_2,\ldots,c_m\), where $e_i$ is the \(i\)-th entity and $c_j$ is the \(j\)-th concept of the entity.

On the entity level, we use \(p_{t_a,i_b}\) to represent the $prediction$ of the concatenation of the \(a\)-th concatenated text and the \(b\)-th image, and \(l_{a,b}\) to represent the label whether it matches. 
Then, we obtain \(L_{entity}\) in a batch:
\begin{equation}
L_{entity} = - \sum_{a=1}^{n} \sum_{b=1}^{n} BCE(l_{a,b}, p_{t_a,i_b})
\end{equation}
where \(BCE\) is the binary cross entropy function.

Similarly, we first obtain concepts related to \(a\)-th entity \(e_a\) using \(C(e_a)\). 
Assuming there are \(m\) concepts of \(e_a\), \(p_{c_k,i_b}\) represents the $prediction$ of the \(k\)-th concept and the \(b\)-th image and \(l_{n \times m \times n}\) represents a matrix where \(l_{a,k,b}\) is 1 if the \(b\)-th entity has the \(k\)-th concept of the \(a\)-th entity; otherwise, \(l_{a,k,b}\) is 0. 
The concept loss \(L_{concept}\) is calculated as:
\begin{align}
& L_{concept}  = \\ \notag 
& - \sum_{a=1}^{n} \sum_{b=1}^{n} \sum_{k=1}^{len(C(e_a))} BCE(l_{a,k,b}, p_{c_k,i_b})
\end{align}

Finally, we update the model parameters by the loss \emph{L} below:
\begin{equation}
L = L_{entity} + L_{concept}
\end{equation}

\subsubsection{Concept-Guided Image-Text Recognition}
\label{method:inference}

\paragraph{\CI}
In \CI, we directly concatenate all concepts $c$ related to the entity $e$ as $t$ and input the concatenated text $t$ and image $i$ into the PVLM. 
For example, take the entity \textit{Jay Chou} associated with concepts like \textit{singer}, \textit{actor}, and \textit{director}. 
The concatenated text would be \textit{Jay Chou, singer, actor, director.}
The PVLM generates a $prediction$ for input \(t,i\) and we set an threshold for judging whether an image-text pair matches.

While \CI improves performance in experiments, it acts as a black-box model lacking explanatory capability. 
Additionally, images of long-tailed entities are scarce. 
The black-box approach's prediction lacks credibility, potentially causing errors or the loss of correct images. 
Therefore, we introduce \EF to re-judge the samples discarded in \CI.

\paragraph{\EF} 
For a more comprehensive understanding of \EF, we first define:

\paragraph{Definition}
$P(·)$ represents the probability of occurrence.
$E$ and $H$ represent the evidence events and the ultimate conclusion, respectively.
$P(E)$ and $P(H)$ are utilized to express the probability of $E$ and $H$.
Additionally, $Con(E, H)$ is the contribution of evidence $E$ on conclusion $H$.

In our task, the evidence \emph{E} refers to the image matching the concept of the entity, while the conclusion \emph{H} is that the image matches the entity.

In \EF, essentially, we transform the task of matching an entity and an image into a comprehensive analysis of the matching between the concepts of the entity and the image. 
In Figure~\ref{fig:framework}, $E_i$ represents an image matching a concept. 
For example, we define evidence $E_1$ as \textit{The object in the image is an animal} and evidence $E_2$ as \textit{The object in the image is an antelope}.
Correspondingly, $H$ can be \textit{The object in the image is Klipspringer}. 
As a result, we directly utilize the prediction of the image and the concept as $P(E)$, where each $E_i$ corresponds to a $P(E_i)$.

The contribution of each evidence $E$ on the conclusion $H$ is different. 
For example, \textit{The object in the image is an antelope} provides more information than \textit{The object in the image is an animal} for judging the image matching \textit{Klipspringer} due to its narrower scope. 
To measure this contribution, we define $Con(E_i, H)$ for each $E_i$ as follows:
\begin{equation} 
Con(E_i,H)=\begin{cases}
\frac{\frac{1}{log(num)} - \frac{1}{ents}}{1 - \frac{1}{ents}} & \text{if } num\geq n \\
1 & \text{if } num<n
\end{cases} \end{equation}
where $num$ denotes the number of entities that contain this concept, 
$ents$ denotes the number of all the entities, and $n$ is the base of $log$ for scaling. (We use 10 in this paper.)
Notably, the contribution is based on the distribution of entities and concepts in the test set, independent of the training process.

Fianlly, the $P(H)$ is calculated as:
\begin{equation}
    P(H) = \frac{1}{n} \sum_{i=1}^{n} P(E_i) \cdot Con(E_i, H)
\end{equation}
In this equation, $n$ denotes the number of concepts of the entity. 
$P(H)$ represents the probability of the conclusion $H$, and we utilize the threshold to determine whether the conclusion $H$ is classified as $True$ or $False$.

\subsection{Human Verification}
Because images of long-tailed entities are rare and valuable, the method also supports human verification to preserve more correct images. 
The scarcity of visual representations for long-tailed entities makes it challenging for annotators to assess the relevance of images directly. 
However, evidence from \EF helps overcome this challenge. 
\method uses \EF to review images discarded in \CI and offer explanations as evidence. 
For example, as shown in Figure~\ref{fig:framework}, it may be difficult for humans to determine if an image depicts a \textit{Klipspringer}. 
However, providing evidence such as \textit{The image matches a mammal} and \textit{The image matches an antelope} greatly assists humans in making more accurate recognition.

\section{Experimental Setup}
\label{sec:Experiments}
\begin{table}[t]
\centering
\small
\begin{tabular}{cc}
\toprule
\textbf{Statistic} & \textbf{Number}\\
\midrule
Total Entities & 25,166 \\
BLC Concepts & 1,278\\
Total Concepts & 10,702\\
Average BLC Concepts per Entity & 2.78 \\
Average Concepts per Entity & 4.45 \\
\bottomrule
\end{tabular}
\caption{Statistics of the dataset.}
\label{tab:statistics}
\end{table}

\subsection{Data Collection}
To address the absence of a suitable dataset for long-tailed image-text recognition, we employ a rule-based method that uses entity linking to accurately identify images of long-tailed entities. 
Using this dataset, we evaluate our method on various downstream tasks and offer a detailed analysis.

Although some long-tailed image classification datasets exist \cite{liu2019large,cui2019class}, they are not suitable for our tasks. 
The long tail of these datasets is designed for model training rather than representing genuine scarcity. 
To tackle this, we select long-tailed entities from an actual KG CN-DBpedia~\cite{xu2017cn}. 
Using entity linking~\cite{chen2018short}, we collect pertinent images for these entities. 
As a result, we obtain a dataset with 25,166 image-text pairs of long-tailed entities and convert these entity names into English.

\paragraph{Selection of Long-Tailed Entities}
To identify long-tailed entities, we analyze the distribution of entities in CN-DBpedia, focusing on a property called \textit{viewtimes} that reflects their click frequency. 
We randomly select 100 entities from the knowledge graph and further examine their \textit{viewtimes}, as shown in Figure~\ref{fig:powerlaw}. 
Our observations reveal a positive correlation between an entity's \textit{viewtimes} and the number of its images. 
Thus, we classify entities with \textit{viewtimes} below 100,000 as long-tailed entities, which typically have few or no images available online.

\begin{figure}[t] 
    \centering
        \includegraphics[width=\linewidth]{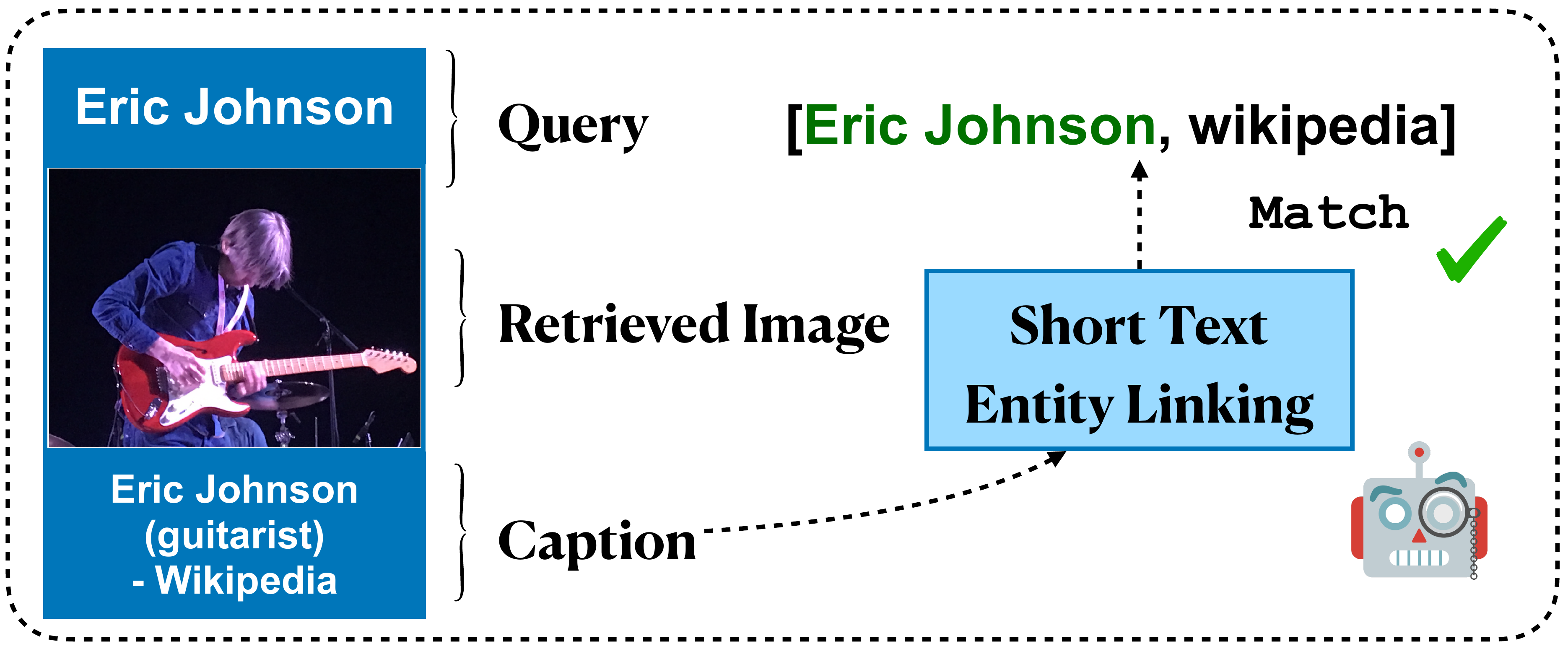}
    \caption{The process of obtaining correct images through short text entity linking.}
    \label{fig:entitylinking}
\end{figure}

\paragraph{Grounding Long-Tailed Entities through Entity Linking}
\begin{table}[t]
\centering
\small
\begin{tabular}{lccc}
\toprule
\textbf{Method} & \textbf{Precision} & \textbf{Recall} & \textbf{F1} \\
\midrule
Entity Linking & 98 & 62 & 75 \\
\bottomrule
\end{tabular}
\caption{The results of the entity linking method for finding matching images.}
\label{tab:entitylinking}
\end{table}
In response to the inaccurate recognition of PVLMs for long-tailed entities, we use the entity linking approach to find correct images, as depicted in Figure~\ref{fig:entitylinking}. 
Initially, we search for entity names using a search engine. 
Following that, we employ short text entity linking~\cite{chen2018short} on the caption of the top search result image to pair it with the target entity. 
If the entity name is included in the linking results, the image is considered a match.
We pick 100 entities with fewer than 100,000 \textit{viewtimes}, search for images through Google, and manually annotate whether the first image corresponds to the entity. 
As shown in Table~\ref{tab:entitylinking}, our strategy yields high precision, allowing us to accurately generate a dataset of image-text pairs of long-tailed entities.

\paragraph{Statistics}
We use CN-Probase~\cite{chen2019cn}, a comprehensive Chinese concept graph, to gather concepts related to entities from CN-DBpedia. Table~\ref{tab:statistics} shows that our dataset contains approximately 10k concepts for 25k entities, averaging 4.45 concepts per entity. 
Although BLC concepts represent a small portion, each entity has about 3 BLC concepts on average, demonstrating the abundance and prevalence of BLC concepts.

\subsection{Settings}

\paragraph{Metrics}
For classification, we evaluate the performance of \method using accuracy, precision, recall, and F1 score. 
For ranking, we use metrics like Mean Reciprocal Rank (MRR), which is the average of the reciprocal ranks of the first correct answer, Mean Rank (MR), the average rank of the correct answer, and Hit@$k$, the percentage of correct candidates in the top-$k$ predictions of the model in the test set. 
We set $k$ to 1, 5, and 10 in our experiments.

\paragraph{Model Choice} 
We conduct \method on three PVLMs, including CLIP~\cite{radford2021learning}, ALIGN~\cite{jia2021scaling}, and BLIP~\cite{li2022blip}.

\paragraph{Datasets for Downstream Tasks}
Firstly, the image-text pairs gathered are separated into training, validation, and test sets using an 8:1:1 ratio. 
This results in 20,132 training, 2,517 validation, and 2,517 test samples. 
Each piece of training data follows the $(entity, image, label)$ structure, with all labels being $True$. 

In terms of ranking, the validation and test sets include samples with one entity and 50 candidate images. 
To conserve computational resources, we choose 49 negative samples randomly, following the methods of~\cite{teru2020inductive,zha2022inductive}; only one image is correct. 

For classification, we add an equal number of negative samples to the validation and test sets by replacing images from different entities. 
As a result, the classification dataset contains 20,132 training samples, and each of the validation and test sets contains 5,034 samples.

\paragraph{Implementation Details}
We perform our experiments with a single RTX3090 GPU. The batch sizes are set to 64 for CLIP~\cite{radford2021learning}, 4 for ALIGN~\cite{jia2021scaling}, and 16 for BLIP~\cite{li2022blip}. We use the AdamW optimizer with a learning rate of 1e-5.

For the classification task, we apply a threshold of 0.5 to decide if an image corresponds to a text, since the $prediction$ varies from 0 to 1.

\section{Results and Analysis}
\label{sec:Analysis}
\subsection{Main Results}
\setlength\tabcolsep{2pt}
\begin{table}[t]
\centering
\small
\begin{tabular}{lccccc}
\toprule
\textbf{Models} & \textbf{MR $\downarrow$} & \textbf{MRR $\uparrow$} & \textbf{Hit@1 $\uparrow$} & \textbf{Hit@5 $\uparrow$} & \textbf{Hit@10 $\uparrow$} \\
\midrule
CLIP & 13.22 & 27.10 & 15.45 & 27.25 & 52.36 \\
 \ \ w/ \textit{Stage1}  & \textbf{5.51} & \textbf{50.14} & \textbf{33.65}& \textbf{58.72} & \textbf{84.51} \\
\midrule
ALIGN & 13.04 & 27.72 & 15.97 & 28.29 & 52.88 \\
 \ \ w/ \textit{Stage1}  & \textbf{5.47} &\textbf{49.81}&\textbf{33.73} & \textbf{57.37} & \textbf{84.74} \\
\midrule
BLIP & 14.21 & 21.09 & 8.34 & 21.37 & 49.30 \\
 \ \ w/ \textit{Stage1} & \textbf{7.04} & \textbf{38.00} & \textbf{19.39} &\textbf{46.60} & \textbf{77.91}\\
\bottomrule
\end{tabular}
\caption{Results for the ranking task. \textit{Stage1} represents \CI in our framework.}
\label{tab:mainex_link}
\end{table}

\setlength\tabcolsep{5.2pt}
\begin{table}[t]
\centering
\small
\begin{tabular}{lcccc}
\toprule
\textbf{Models} & \textbf{Accuracy} & \textbf{Precision} & \textbf{Recall} & \textbf{F1}\\
\midrule
CLIP &  67.44  & 62.37 & 88.37  & 73.13 \\
 \ \ w/ \textit{Stage1}  & 83.63 & 81.67 & 87.10 & 84.30 \\
 \ \ w/ \textit{Stage1+2} & \textbf{83.87} & 80.92 & 88.64 & \textbf{84.60} \\
\midrule
ALIGN  & 68.12 & 63.12 & 89.38 &  73.99 \\
 \ \ w/ \textit{Stage1}  & \textbf{83.19} & 77.82 & 92.84 &  84.67 \\
 \ \ w/ \textit{Stage1+2} & 83.13 & 77.84 & 92.67 & \textbf{84.68} \\
\midrule
BLIP  & 68.55 & 61.58 & 91.30 &71.30 \\
 \ \ w/ \textit{Stage1}  & 79.41 & 76.61 & 84.70 & 80.45 \\
 \ \ w/ \textit{Stage1+2} & \textbf{79.42} & 76.42 & 85.10 & \textbf{80.53} \\
\bottomrule
    \end{tabular}
\caption{Results for the classification task.
\textit{Stage1} and \textit{Stage2} repersents \CI and \EF in our framework respectively.}
\label{tab:mainex}
\end{table}

Table~\ref{tab:mainex_link} compares the performance with and without concept guidance in the ranking task. 
Incorporating \CI (\textit{Stage1}) significantly enhances the performance of all PVLMs, highlighting the importance of integrating concepts.
Table~\ref{tab:mainex} shows performance in the classification task under three conditions: without concepts, with only \CI, and with both \CI and \EF. 
\CI considerably improves recognition accuracy through the appropriate integration of concepts. 
Additionally, \EF, aimed at providing explainability, further enhances performance.

Experimental results show that integrating concepts allows PVLMs to more effectively align image and text modalities. 
PVLMs associate images with various concepts related to entities, not just the names of entities, during pre-training. 
\CI improves the recall of knowledge gained in pre-training. 
However, relying solely on this method is inadequate due to its black-box nature. 
As a result, we introduce the \EF module, which breaks down the recognition process into multiple pieces of evidence for the conclusion. 
This decomposition maintains effective performance while also making it more convincing and explicit.

Notably, Tables~\ref{tab:mainex_link} and~\ref{tab:mainex} demonstrate the superior performance of our method across different models, showcasing its flexibility. 
All models can determine whether a piece of text and an image match are suitable for integration into our method. 
Therefore, our method is model-pluggable and supports the replacement of different PVLMs or other methods used for image-text matching, significantly enhancing its transferability.

\subsection{Analysis}
\label{exp:analysis}
\begin{figure}[t]
    \centering
        \includegraphics[width=1.0\linewidth]{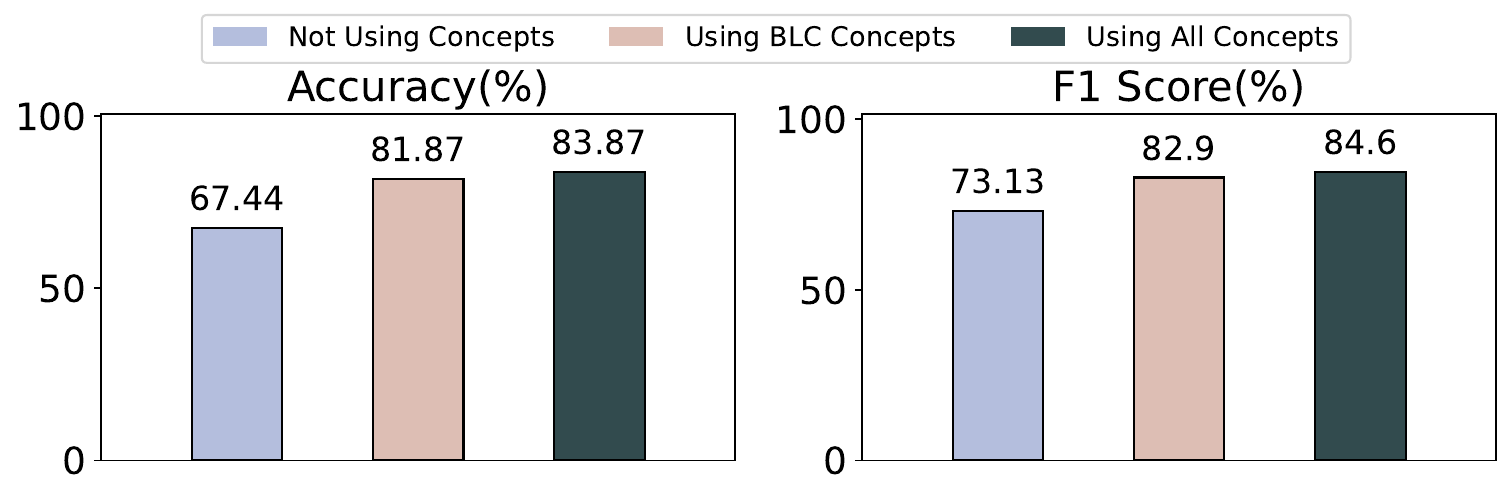} 
    \caption{Comparison of using different concepts in our framework. \textit{Not Using Concepts} represents using only entity
names. \textit{Using BLC Concepts} and \textit{Using All Concepts} represents using BLC and all concepts respectively.}
    \label{fig:conceptselection}
\end{figure}

\paragraph{How do different concept selections affect performance?}
To explore how concept selection affects results, we employ the same method but with different concepts. 
Figure~\ref{fig:conceptselection} shows the influence of using various concepts on the classification task. 
It indicates that using Basic-level Categorization (BLC) concepts improves performance, but incorporating all concepts leads to optimal outcomes. 
This suggests that BLC concepts are effective for identifying unfamiliar entities, and fine-grained concepts are also crucial as PVLMs can utilize detailed concept knowledge. 
Thus, we decide to include all concepts in our main experiments. 
However, Table~\ref{tab:statistics} and Figure~\ref{fig:conceptselection} also demonstrate that BLC concepts, being more common, still offer competitive performance. 
Considering that fine-grained concepts may be rare and difficult to collect, as highlighted by some studies~\cite{li2021towards,yuan-etal-2022-generative-entity,yuan-etal-2023-causality}, BLC concepts present a practical alternative.

\setlength\tabcolsep{3pt}
\begin{table}[t]
\centering
\small
\begin{tabular}{lcccc}
\toprule
\textbf{Methods} & \textbf{Accuracy} & \textbf{Precision} & \textbf{Recall} & \textbf{F1} \\
\midrule
\textit{Search Engine}  & 61.11 & 61.11 & 100 & 75.77  \\
CLIP & 73.33 & 73.13 & 89.09 & 80.33 \\
\ \  w/ \textit{Stage1+2} & \textbf{81.11} & 82.76 & 87.27 & \textbf{84.96} \\
\bottomrule
\end{tabular}
\caption{
\textit{Search Engine} indicates entity grounding via a search engine. CLIP represents not using concepts. \textit{Stage1} and \textit{Stage2} repersents \CI and \EF in our framework respectively.}
\label{tab:baseline}
\end{table}

\paragraph{How does the performance of the method measure up against traditional approaches?} 
To show the comparison, we select 100 long-tailed entities and annotate whether the first searched image and the entity match. 
We then employ a trained CLIP model, using only entity names and our concept-guided framework \method. 
As shown in Figure~\ref{tab:baseline}, conventional methods that depend on direct search engine may attain a 100\% recall rate, but suffer from low precision, resulting in unsatisfactory F1 scores. 
On the other hand, using CLIP demonstrates enhancements, stressing the importance of PVLMs in entity grounding. 
Our \method further amplifies the performance of PVLMs by incorporating concepts, underlining the method's efficiency and the significant advantage gained from integrating concepts.

\begin{figure}[b]
    \centering
        \includegraphics[width=1.0\linewidth]{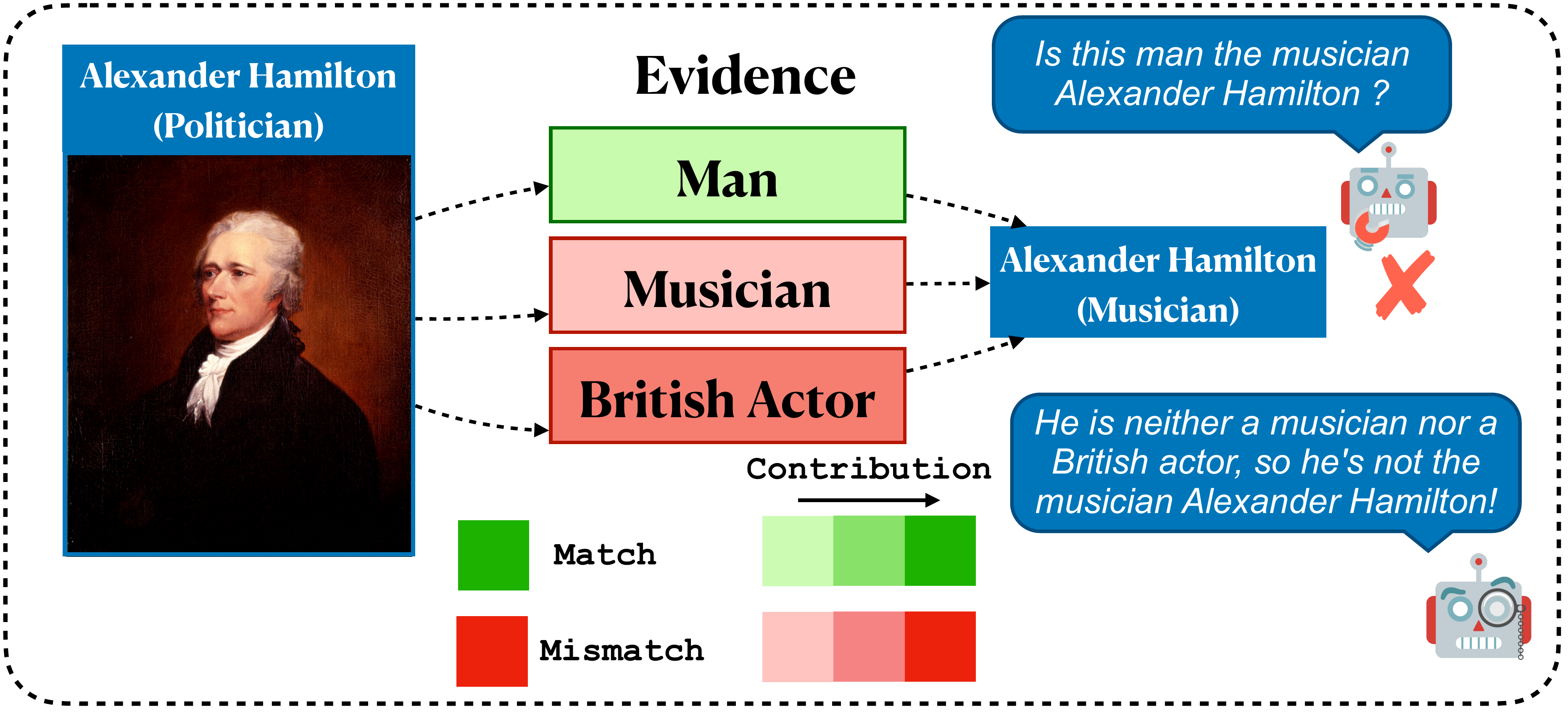}
    \caption{The process of recognizing a long-tailed entity with evidence.
    }
    \label{fig:explain}
\end{figure}

\setlength\tabcolsep{3pt}
\begin{table}[t]
\centering
\small
\begin{tabular}{lcccc}
\toprule
\textbf{Methods} & \textbf{Accuracy} & \textbf{Precision} & \textbf{Recall} & \textbf{F1} \\
\midrule
\textit{\method} & 80.00 & 76.78 & 86.00 & 81.13 \\
\textit{Human Verification}  & 75.00 & 68.38 & 93.00 & 78.81  \\
\ \ \ + \textit{Evidence}  & \textbf{83.00} & 77.50 & 93.00 & \textbf{84.54} \\
\bottomrule
\end{tabular}
\caption{\method shows the performance of our two-stage framework. \textit{Human Verification} shows results from \method with human verification, and \textit{Evidence} denotes human verification with extra evidence from \EF.}
\label{tab:humanannotate}
\end{table}

\paragraph{How does \EF support human verification?}
Due to the scarcity of images of less common entities, we choose not to discard images labeled as incorrect by \CI, but rather to use \EF for re-judging. 
In Figure~\ref{fig:explain}, two entities are named \textit{Alexander Hamilton}. 
When the goal is to find an image of the musician \textit{Alexander Hamilton} but an image of the politician with the same name is retrieved instead, \EF helps clarify this error. 
It shows that the evidence \textit{The person in the image is a man} is true, but \textit{The person in the image is a musician} and \textit{The person in the image is an English actor} are false. 
This evidence explains why the image does not match the musician \textit{Alexander Hamilton}, aiding in the identification of mismatches and supporting human annotators in their verification work, especially for less common entities where direct judgment is challenging.

To further investigate the role of evidence in human annotation, we conduct a test with 200 image-text pairs. 
These are evenly divided into positive and negative samples, and a two-stage classification method is applied. 
For samples identified as mismatches, we engage five annotators to review these mismatches. 
The accuracy and F1 scores are recalculated after this annotation. 
As indicated in Table~\ref{tab:humanannotate}, our findings demonstrate that explainability significantly enhances the verification process, highlighting the value of evidence in recognizing unfamiliar entities.

\begin{figure}[t]
    \centering
        \includegraphics[width=\linewidth]{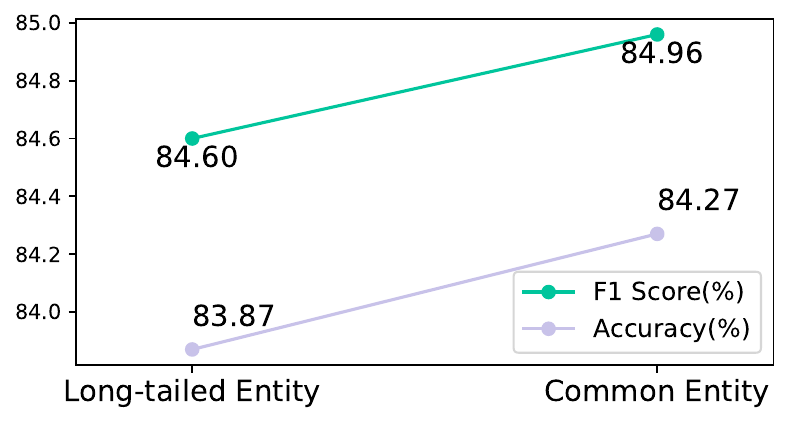} 
    \caption{Comparison of our method on different entities.}
    \label{fig:common}
\end{figure}

\paragraph{Is our method general and robust?}
We randomly select 100 common entities (with \textit{viewtimes} over 1,000,000, in contrast to long-tailed entities with \textit{viewtimes} under 100,000), using the best checkpoint trained in Table~\ref{tab:mainex}. 
As illustrated in Figure~\ref{fig:common}, \method is beneficial for both long-tailed and common entities. 
This proves that concept guidance brings a generally effective enhancement, aligning with human cognition.

Our framework gains robustness through concept aggregation. 
Incorporating concepts is beneficial but may introduce noise, particularly when PVLMs struggle with fine-grained concepts. 
The more detailed a concept is, the closer it is to the entity, which complicates recognition. 
We mitigate noise by aggregating multiple concepts.
During \CI, we concatenate all the concepts, and in \EF, we synthesize all evidence to draw conclusions. 
In both modules, we find that fine-grained concepts offer improvements over using only BLC concepts, indicating that aggregation effectively reduces noise.

\section{Conclusion}
\label{sec:Conclusion}
We propose a two-stage framework \method, using PVLMs with concept guidance to ground long-tailed entities in MMKGs effectively. 
Our experimental results demonstrate that \method greatly improves the ability of PVLMs to recognize image-text pairs of long-tailed entities. 
Furthermore, recognizing unfamiliar entities through concepts is convincing and provides clear evidence for human verification, suggesting a future direction for better handling long-tailed entity grounding.

\section*{Limitation}
\label{sec:Limitation}
Throughout our method, we utilize concepts from CN-Probase, which contains noise.
Both the quantity and quality of these concepts play a crucial role in determining the performance of our method. 
Exploring alternative concept generation methods can serve as a potential research question for future research.
The improvement of concepts in the future is expected to contribute to the enhancement of our methods for more accurate long-tailed entity grounding.

\section*{Ethical Considerations}
\label{sec:Ethical}
We hereby acknowledge that all authors of this work are aware of the provided ACL Code of Ethics and honor the code of conduct.

\paragraph{Use of Human Annotations} 
All raters have been paid above the local minimum wage and consented to use the evaluation dataset for research purposes in our paper. 
Human annotations are only utilized in the methodological research stages to assess the proposed solution's feasibility. 
To guarantee the security of all annotators throughout the annotation process, they are justly remunerated according to local standards.

\paragraph{Risks} 
The datasets used in this paper are obtained from public sources and anonymized to protect against any offensive information.

\section*{Acknowledgements}
We thank the anonymous reviewers for their valuable comments.
This work is supported by National Natural Science Foundation of China (No. 62102095), and Science and Technology Commission of Shanghai Municipality Grant (No. 22511105902).
The computations in this research are performed using the CFFF platform of Fudan University.

\bibliography{anthology}
\bibstyle{acl_natbib}

\appendix



\end{document}